\documentclass{article}
\usepackage{arxiv}

\usepackage[utf8]{inputenc} 
\usepackage[T1]{fontenc}    
\usepackage{amsmath}      
\usepackage[hyphens]{url}        
\usepackage{graphicx}
\RequirePackage{filecontents}    
\usepackage{booktabs}       
\usepackage{amsfonts} 
\usepackage{booktabs}
\usepackage{nicefrac}       
\usepackage{microtype}      
\usepackage{lipsum}
\usepackage{fancyhdr}       
\usepackage[table]{xcolor}
\usepackage{authblk}
\usepackage{doi}
\usepackage[numbers]{natbib}
\usepackage{hyperref}
\usepackage{cleveref}
\usepackage{ragged2e}
\usepackage{url}

\title{Early prediction of onset of sepsis in Clinical Setting}

\renewcommand * {\Authfont}{\bfseries}
\author{\thanks{corresponding author}
	{	Fahim Mohammad\textsuperscript{1}, 
		Lakshmi Arunachalam\textsuperscript{1}, 
		Samanway Sadhu\textsuperscript{1}, 
		Boudewijn Aasman\textsuperscript{2}, 
		Shweta Garg\textsuperscript{3}, 
		Adil Ahmed\textsuperscript{3},
		Silvie Colman\textsuperscript{2}, 
		Meena Arunachalam\thanks{The author contributed when she was as an employee of  Intel\textsuperscript{\textregistered}. Currently at AMD, Santa Clara.},
		Sudhir Kulkarni\textsuperscript{1},
		Parsa Mirhaji\textsuperscript{3}
}

\textit{\bf{email:} } \texttt{fahim.mohammad, lakshmi.arunachalam, samanway.sadhu@intel.com, baasman@montefiore.org, shweta.garg, adil.ahmed@einsteinmed.edu, scolman@montefiore.org, meena\_aruna@yahoo.com, sudhir.kulkarni@intel.com, parsa.mirhaji@einsteinmed.edu}
\newline
\\
\textsuperscript{1} Intel\textsuperscript{\textregistered} Corporation, Hillsboro, OR, USA
\\
\textsuperscript{2} Montefiore Medical Center, Bronx, NY, USA
\\
\textsuperscript{3} Albert Einstein College of Medicine, Bronx, NY, USA
}

\renewcommand{\shorttitle}{Early prediction of onset of sepsis in Clinical Setting}

\hypersetup{
	pdftitle={Early prediction of onset of sepsis in Clinical Setting},
	pdfsubject={cs.AI, cs.CV, cs.CL },
	pdfauthor={Fahim Mohammad, Lakshmi Arunachalam, Samanway Sadhu, Boudewijn Aasman, Shweta Garg, Adil Ahmed, Meena Arunachalam, Sudhir Kulkarni, Parsa Mirhaji},
		pdfkeywords={Sepsis Detection, AI applications, Electronic Health Records}}

\begin{document}
	\maketitle
	\begin{abstract}
		This study proposes the use of Machine Learning models to predict the early onset of sepsis using deidentified clinical data from Montefiore Medical Center in Bronx, NY, USA. A supervised learning approach was adopted, wherein an XGBoost model was trained utilizing 80\% of the train dataset, encompassing 107 features (including the original and derived features). Subsequently, the model was evaluated on the remaining 20\% of the test data. The model was validated on prospective data that was entirely unseen during the training phase. To assess the model's performance at the individual patient level and timeliness of the prediction, a normalized utility score was employed, a widely recognized scoring methodology for sepsis detection, as outlined in the PhysioNet Sepsis Challenge paper. Metrics such as F1 Score, Sensitivity, Specificity, and Flag Rate were also devised. The model achieved a normalized utility score of 0.494 on test data and 0.378 on prospective data at threshold 0.3. The F1 scores were 80.8\% and 67.1\% respectively for the test data and the prospective data for the same threshold, highlighting its potential to be integrated into clinical decision-making processes effectively. These results bear testament to the model's robust predictive capabilities and its potential to substantially impact clinical decision-making processes. 
	\end{abstract}


\section{Background}

		Sepsis is defined as a life-threatening organ dysfunction caused by a dysregulated host response to infection \cite{singer2016third}. In normal cases, the body releases chemicals to fight infection. But when sepsis occurs, organs may react differently to these chemicals causing multiple organ failures eventually resulting in death. Early recognition of sepsis is crucial to reduce mortality rates, and administering antibiotics promptly is vital for patient recovery. However, there is currently no single specific test for sepsis detection, so doctors rely on monitoring vital signs, collecting biomarker values and alarming clinical signs like increased heart rate, shortness of breath etc. All the information needs to be gathered and analyzed to define if a patient is septic or non-septic.
		Various methodologies have been used in the past for predicting early-onset sepsis. In one online challenge, organized by PhysioNet in 2019 \cite{reyna2020early}\cite{goldberger2000physionet}\cite{physio-online}, researchers employed variety of techniques, including traditional feature engineering, gradient boosting classifiers, XGBoost \cite{chen2016xgboost}  ensembles with sliding window features, and other deep learning approaches. Morrill et al \cite{morrill2019signature} achieved promising results by employing Signature features and hand-crafted features, such as shock Index, BUN/CR, Partial SOFA, and SOFA deterioration, to predict sepsis. They emerged as the top performer in the challenge. Du et al \cite{du2019automated} proposed a gradient boosting classifier algorithm that utilized comprehensive sets of vital signs, lab features, missing value flags, and changes/variance of vital sign variables. Weighted cross-entropy was chosen as the loss function for their model. Zabihi et al \cite{zabihi2019sepsis} developed an ensemble of XGBoost models, incorporating sliding window and non-sliding window-based features. To handle missing data, they employed linear interpolation, achieving improved imputation accuracy. Li et al \cite{li2019tasp} introduced a Time-Phased Approach for Sepsis Prediction (TASP), utilizing distinct sub-models for different ICU stages. Their ensemble involved two tree-based methods (LightGBM) and an RNN-based method, each trained on specific data portions with tailored feature sets. Singh et al.  \cite{singh2019utilizing} used the information missingness along with the XGBoost algorithm and several other variations to achieve good performance. 
			
		NN based models: Several researchers explored deep learning methodologies. He et al \cite{he2019early} leveraged a deep feature extractor using pre-trained LSTM \cite{hochreiter1997long} and combined it with SOFA and SIRS features, using XGBoost and Gradient Boost Decision Tree models for prediction. Lee et al \cite{lee2019improving} introduced a decay-to-default imputation technique for handling missing values. They trained an ensemble of Transformers and LSTM models with auxiliary loss for regularization purposes. Roussel et al \cite{roussel2019recurrent} proposed an embedding module to create compact feature representations. They trained an LSTM model with auxiliary loss to ensure better generalization. Chang et al \cite{chang2019multi} introduced the TCN model and combined it with the RITS imputation technique to address missing data challenges. Nejedly et al \cite{nejedly2019prediction} utilized LSTM models and performed hyperparameter optimization to enhance predictive performance. These research works highlight the diverse approaches taken by the scientific community to predict sepsis onset, leveraging a mix of traditional machine learning, ensemble techniques, and deep learning models to achieve accurate and reliable predictions. The Table \ref{tab:review_method} and Table \ref{tab:review_feature} summarizes the techniques incorporated by above studies. 
			
		In Table 1, the five top ranked papers (paper 1 through paper 5) and their corresponding Normalized utility scores on train and test data is shown along with data imputation techniques, modeling techniques, evaluation techniques and loss functions employed by the authors. Paper 6 through paper 10 are mostly based on Neural Network models such as RNN, LSTM or TCN. Table 2 summarized the feature engineering methods used by those papers. 
							
						\begin{table}
							\centering
							\includegraphics[width=\textwidth]{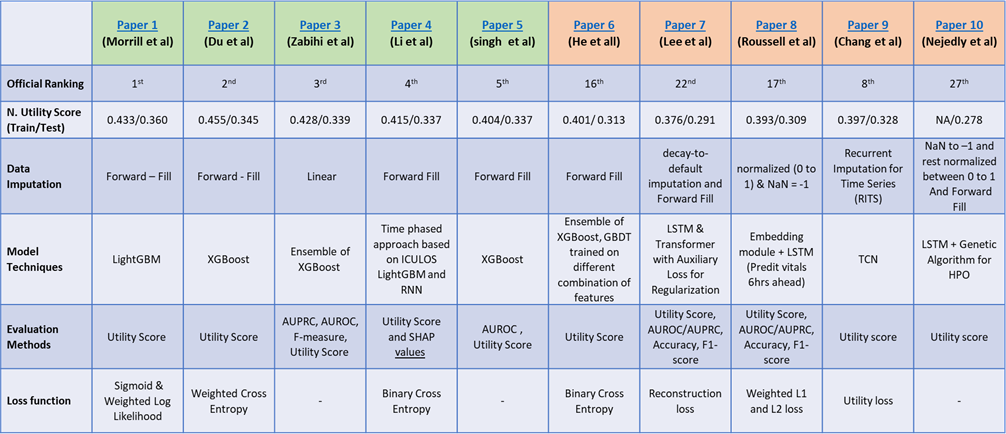}
							\caption{Modeling Techniques for Sepsis Prediction. First five papers are the top ranked paper mostly based on tree based boosting algorithms. Paper 6 through 10 employ the Neural Network based models. "-"  = not known / NA}
							\label{tab:review_method}
						\end{table}

						\begin{table}
							\centering
							\includegraphics[width=\textwidth]{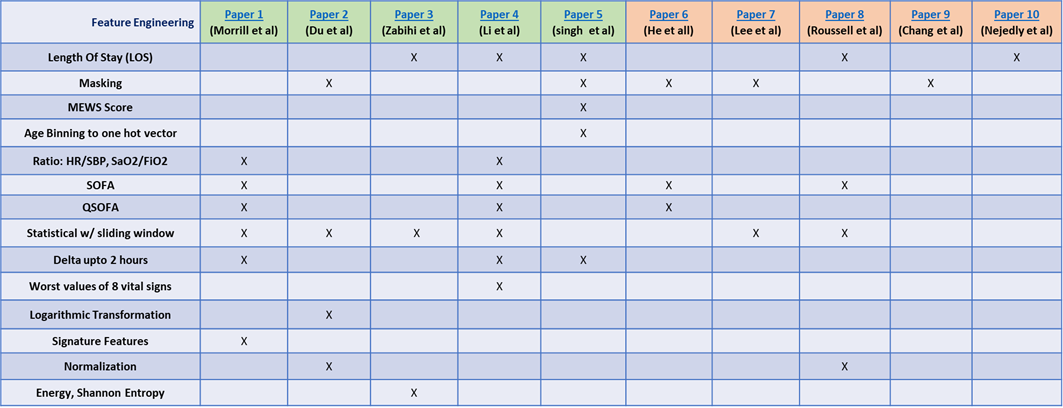}
							\caption{Feature engineering Techniques used for Sepsis Prediction}
							\label{tab:review_feature}
						\end{table}

			In our current work, we propose the utilization of Machine Learning models to detect sepsis up to 6 hours prior to its onset, leveraging the wealth of clinical data collected at Montefiore Medical Center at Bronx, NY, USA. Based on our review of multiple papers mentioned above, we tried to adopt some of the best features and techniques from those papers. As most of the top-ranked papers used the XGBoost model in some form or other, we used XGBoost to predict the likelihood of a patient developing sepsis within a critical time window. The model was trained on a total of 107 features. We tested the model on 20\% of the test data. We also ran an evaluation with prospective data collected by the hospital (completely unseen by the model) at later time point. 
			The following sections outlines the datasets, our method, presents our findings, and offers insights into the implications of early sepsis detection through machine learning.

\section{Methods}
	\label{sec:method}
	Our objective is to predict if a patient is septic in timely manner at any time point \textit{t}. To accomplish the objective, we first clean the data, drop unwanted rows and columns, create a mask to capture the missingness in data, impute the data, add time-based statistical features and rule-based generated features to the data. We split the data with 80-20 ratio into train and test subset, train the model until convergence with XGBoost algorithm and test the model on the 20\% of retrospective data using metrics like normalized utility score, F1-score etc. We finally test the robustness of model on the prospective data, completely unseen by the model. A high-level architechture diagram is seen in the Figure \ref{fig:architecture}.

			 \begin{figure}
			 	\centering
			 	\includegraphics[width=\textwidth]{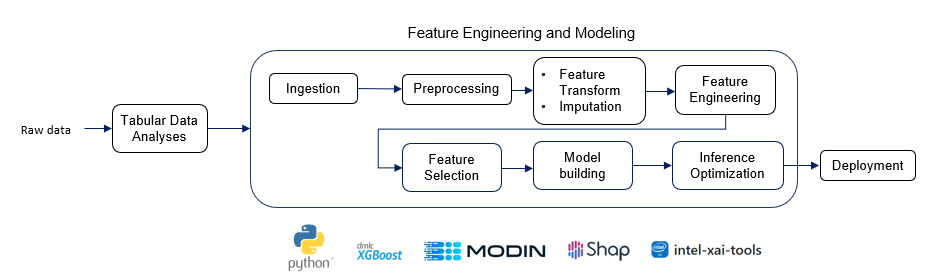}
			 	\caption{A high-level architecture diagram showing steps.}
			 	\label{fig:LOS}
			 \end{figure}

		\subsection{Dataset}
		Our tabular dataset contains data from 1,039,810 encounters. One encounter (represented as an integer ID) corresponds to a patient’s entire stay at the hospital. If a patient is discharged and comes later for hospital admission, then that patient is assigned another encounter ID. Demographics data, vitals and lab results are collected during the entire stay of the patients.  These encounters and the associated data are then converted into hourly data, getting a total of 41,355,296 records. The initial data has a total of 85 features shown in Table \ref{tab:origfeatures}.Most of the lab results and vitals in hourly data has min- and max- components that measure the minimum value and maximum value in that hourly data. 
		
						\begin{table}
							\centering
							\includegraphics[width=\textwidth]{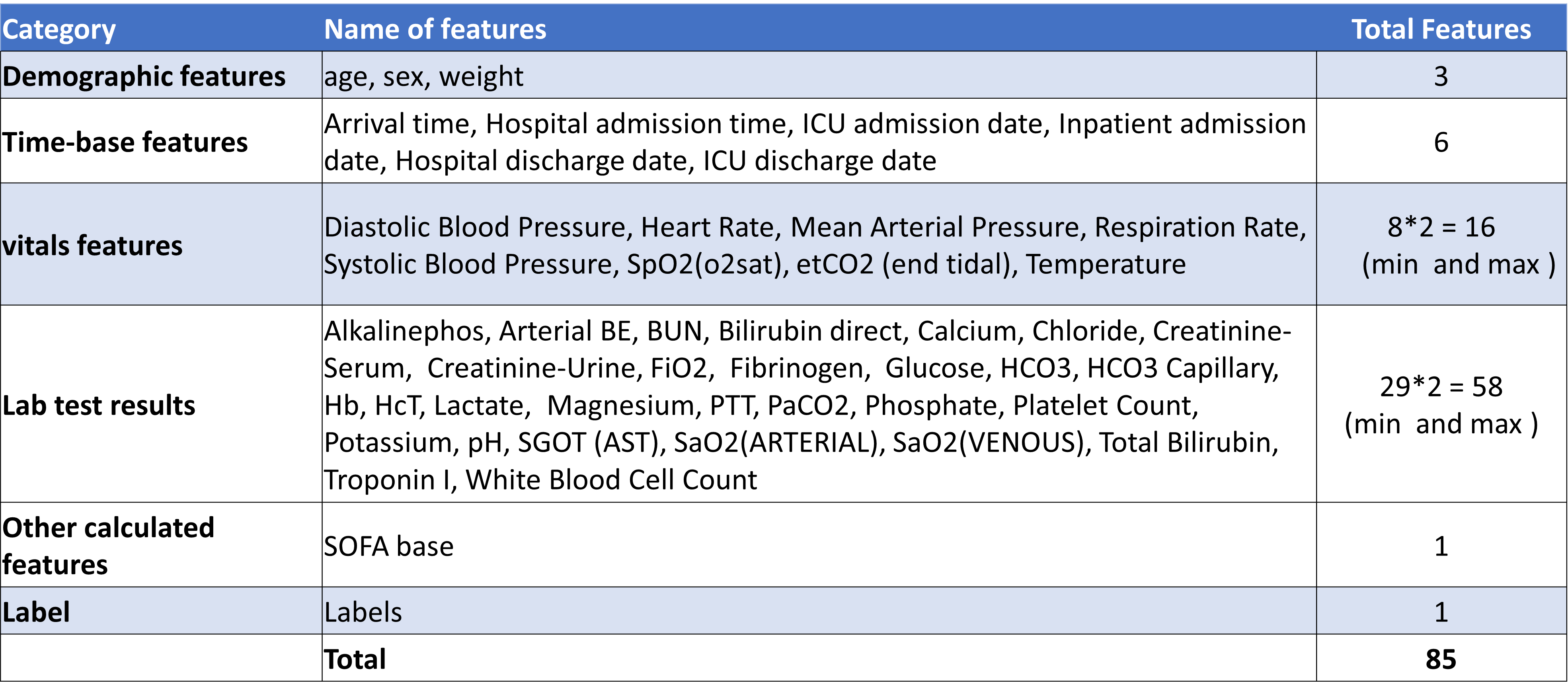}
							\caption{List of original features in raw data and their count}
							\label{tab:origfeatures}
						\end{table}
					
		Typically, the Electronic Health    Records (EHR) data is noisy, and our first task was to clean the data. This involves renaming some of the columns (Column names have been tidied up to drop spaces between them). The demographic features include encountered, age, sex, and weight. One of the features in earlier studies are shown to be length of stay (LOS) at the hospital or ICU \cite{zabihi2019sepsis} \cite{li2019tasp} \cite{singh2019utilizing}. We calculated LOS by finding the difference between the current timestamp and the hospital admission time. All other time features were dropped. Many vital features and lab test results have a mix of hourly min- and max- values. When multiple readings are taken within an hour, both min- and max- values are recorded. Often, the min- and max- values are the same because these values are not continuously sampled every hour. We measured 8 vitals including Temperature, Heart Rate, Respiration Rate, Systolic Blood Pressure, Mean Arterial Pressure, Diastolic Blood Pressure, SpO2(O2sat) and etCO2 (end tidal). We have min- and max- version of the above vitals making the total number of features to be 16.   We are also considering a total of 29 lab test results (58 in total). Additionally, we have the SOFA (Sequential Organ Failure Assessment) score calculated from the raw data. Some post processing features are also included for downstream performance analyses. As we are trying to predict the sepsis 6 hours in advance, all the labels are time-shifted by 6 hours for encouraging early detection of sepsis. 

		\subsection{Data Cleaning}
		We started with a total of 41,355,296 hourly records. During cleanup we dropped records of patients who stayed less than five hours as we may not have much data for the patient. Most of the patients that are admitted remain in hospital for only few days. However, we do have some patients who stayed in the hospital for several months.  To remove some of the outliers, we looked at the percentile values. A 99.99 percentile is close to 695 hours in our data. We round this value to 700 hours which is close to one month of data. We also removed records for patients 3 days after their discharge, some of the test results may arrive after discharge of the patient. We dropped patients older than 105 years old. All the above-mentioned cleaning gave us a total of 36,968,285 records.  The Length Of Stay in hours (LOS) is skewed and its distribution can be seen in the Figure \ref{fig:LOS}.

				\begin{figure}
					\centering
					\includegraphics[width=\textwidth]{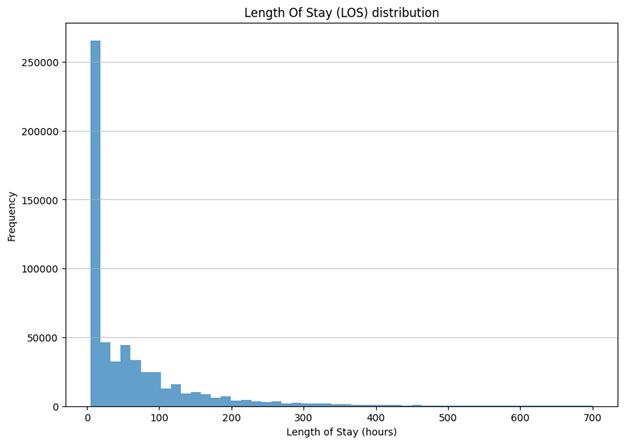}
					\caption{Length of Stay (LOS) of patients in the Hospital}
					\label{fig:LOS}
				\end{figure}
			
		\subsection{Removing highly collinear features}
			The vitals and lab test values contains the min\_and max\_components, which, most of the time are almost the same. The vitals have a total of 16 (8*2) features while the lab tests have 58 (29*2) features. We systematically remove the highly correlated features from the set of vitals and lab test feature by first finding the correlation matrix which is later converted into distance matrix. The distance matrix is then converted into linkage matrix and then dendrogram is constructed using the hierarchical clustering. Using a Ward distance of 1, we found one representative feature from each cluster. In the end, we got a total of 35 final features from an initial list of a total of 74 vitals and lab test features. 

				\begin{figure}
					\centering
					\includegraphics[width=\textwidth]{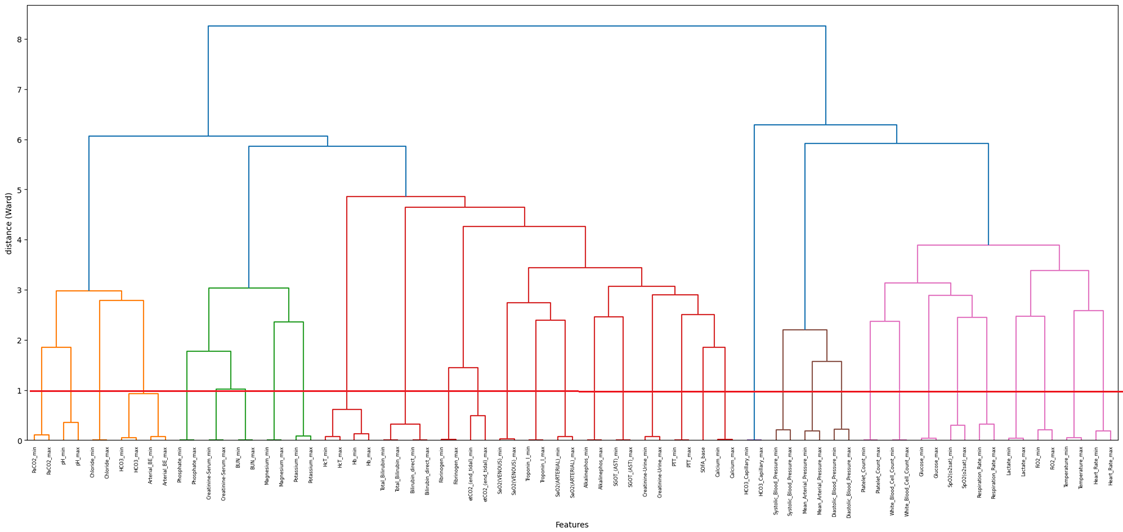}
					\caption{Using dendrogram and ward distance to find representative features from each clusters.}
					\label{fig:Dendrogram}
				\end{figure}
	
		The dataset is highly imbalanced with 5.79\% of all patients have sepsis. Finally for each vitals and lab feature, we provide basic statistics in the Table \ref{tab:featurestats}.

				\begin{table}
					\centering
					\includegraphics[width=\textwidth]{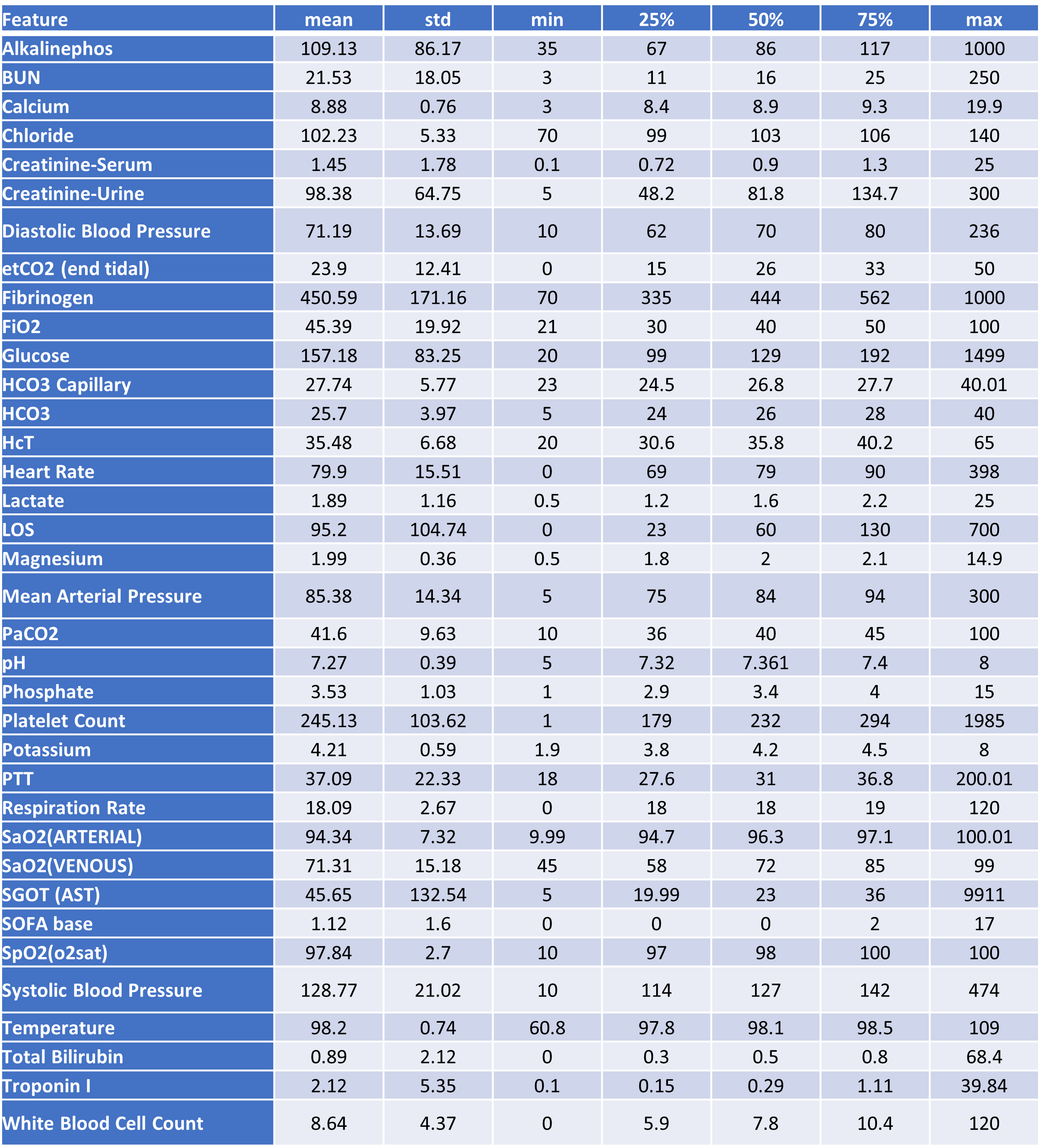}
					\caption{Basic statistics of the features}
					\label{tab:featurestats}
				\end{table}
					
		\subsection{Missingness of data and Masking}
		The vital features have around 20\% of missing data. The lab features are sparsely populated with some of them having more than 90\% of data missing. Dropping these columns is not a viable option since lab measurements are not taken hourly, and some measurements may not be recorded at all during a patient's entire stay, which is expected. To handle the missing data, a masking technique was employed for both vital signs and lab measurements. This process involved creating a binary mask, where present values were marked as 1, and missing values (NaN) were marked as 0. These masked features retains valuable information about the frequency of data collection and may help identify patterns that may be indicative of particular conditions or factors relevant to prognosis (in our case, patient's risk of sepsis).  The Masked features can also provide additional information that might not be explicitly captured by the original features. In cases where the presence or absence of specific values in certain features is clinically relevant, the masked features can contribute valuable information to the overall predictive model.
		
		After applying the masking, the missing data was imputed on per patient basis. Two techniques were explored: forward fill and linear interpolation. Forward fill was utilized to impute any missing demographic values. However, based on experimental results, it was found that linear interpolation was a more effective method for estimating missing values in lab features and vital signs. Linear interpolation utilizes a continuous trend between observed data points to make predictions for missing values, preserving the underlying patterns and trends in the data. By combining masking and linear interpolation, we successfully imputed the missing data, making the dataset complete and more suitable for training the predictive model to identify early-onset sepsis effectively.

		\subsection{Feature Engineering}
		We performed sepsis prediction using four standard sepsis-related scoring systems: Sequential Organ Failure Assessment (SOFA)\cite{guirgis2018updates}, quick Sequential Organ Failure Assessment (qSOFA)\cite{small2020qsofa}, Modified Early Warning Score (MEWS)\cite{mruk2021interobserver}, and Systemic Inflammatory Response Syndrome (SIRS)\cite{small2020qsofa}. These scoring systems were calculated based on various lab values and vital signs, such as Temperature, Respiratory Rate (Resp), Mean Arterial Pressure (MAP), Bilirubin, and Systolic Blood Pressure (SBP).
		
		To further enhance the predictive capability of the model we also calculated additional indicators, including the Shock Index (Heart Rate divided by Systolic Blood Pressure), the ratio of Blood Urea Nitrogen to Creatinine, and the ratio of Oxygen Saturation (SaO2) to Fraction of Inspired Oxygen (FiO2).
		
		As the ICU Length of Stay was identified as an important feature for sepsis prediction in a related study using PhysioNet data [3]. We calculated this feature for their dataset by deriving the ICU Length of Stay by computing the difference between the ICU admission time and the ICU discharge time for each patient.
		
		To incorporate temporal information and better capture the patient's condition over time, we created sliding windows of the last 6 hours for both the lab values and vital signs. These sliding windows were used as input to their classifier, allowing the model to consider recent trends and changes in the patient's condition.
		
		Additionally, we calculated the differences (deltas) in lab values and vital signs between the current hour and the previous hour. This approach helped assess the stability of a patient's condition by measuring changes between consecutive time points.
		
		Furthermore, we also computed various statistical features, such as the mean, minimum, maximum, median, variance, energy, and slope of the lab values and vital signs within the last 6-hour window. These statistical features provided additional insights into the patient's physiological state and variation over the specified period. 
		
		Overall, the combination of standard scoring systems, additional indicators, ICU Length of Stay, sliding windows, delta values, and statistical features aimed to improve the accuracy and effectiveness of sepsis prediction in the study. A list of the definition of generated features are presented in Table \ref{tab:featuredef}
		
						\begin{table}
							\centering
							\includegraphics[width=\textwidth]{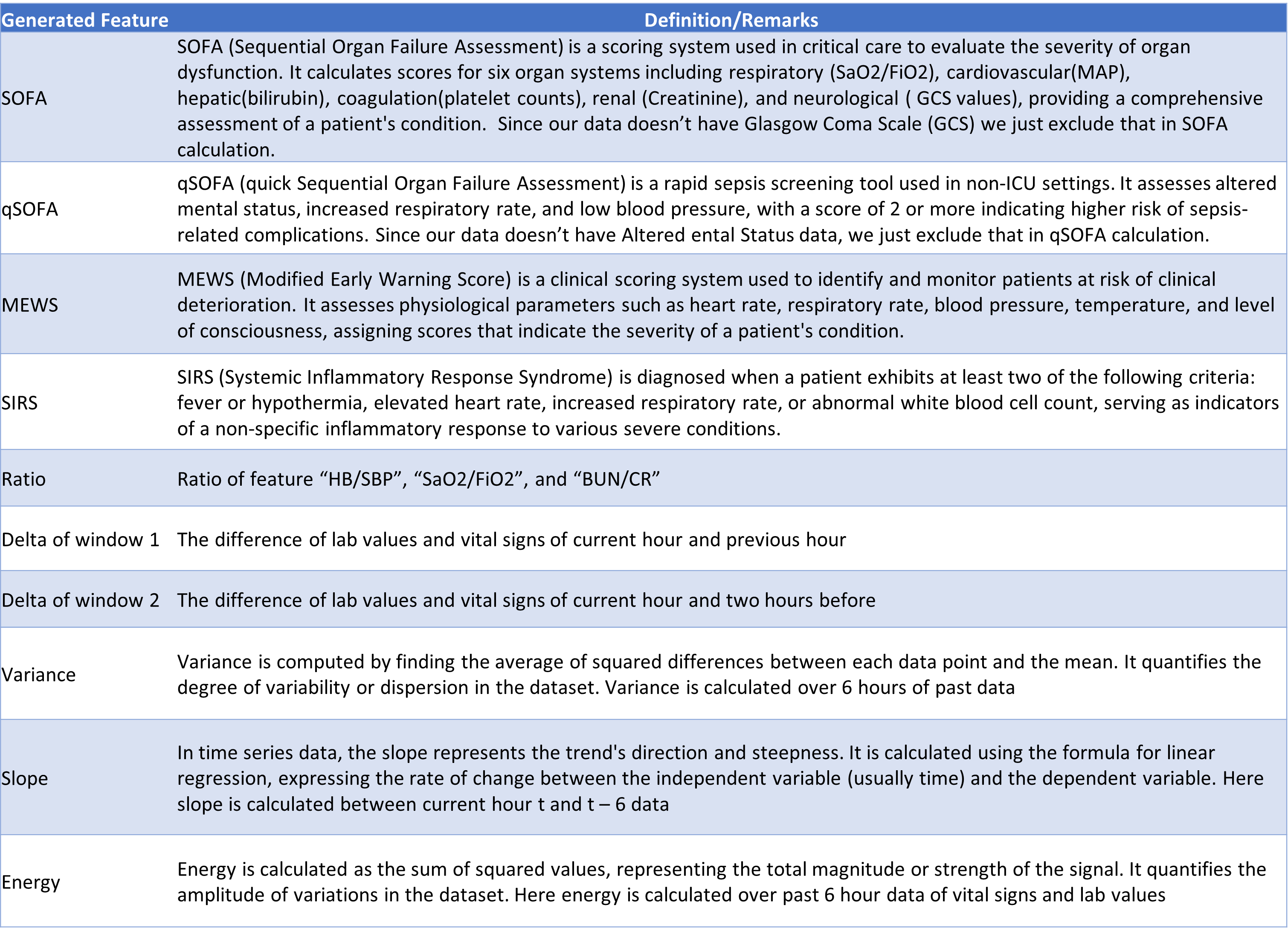}
							\caption{Definition of some of the derived features}
							\label{tab:featuredef}
						\end{table}

		\subsection{Data Preparation Steps}
			Following are the steps taken to get the final features from raw data: 
				\begin{itemize}
					\item 
						Clean the raw data by renaming columns, removing the features that will not be used and cleaning data by removing outliers or removing patients having insufficient data as describe in preprocessing section.  
					\item	
						Remove highly correlated features explained in the above section. In this step, we started with 74 vitals and lab results features and got 35 features (aka original features) for downstream analyses. 
					\item 
						Generate masked features: Use each of these 35 original features to calculate masked features resulting in 35 additional features. These masked features are binary features and will not have any missing values. 
					\item 
						Perform linear interpolation to fill missing values on the original features. Now the original features will be imputed with the data. Some of the entries will still be missing, such as some lab results whose values are unknown till time point \textit{t}. For patient a lab value is never recorded will have missing entries.
					\item 
						Generation and selection of statistical features: We calculate the statistical features using including Delta1 (the difference between the current hour and the previous hour), Delta2 (difference between the current hour and two hours before), variance, slope and energy on each of the imputed original features resulting in 175 derived features based on statistical calculations: 35 (delta1) + 35 (delta2) + 35 (variance) + 35 (slope) + 35(Energy) = 175 features. These statistical features are based on time windows to capture the time components in the data. The Variance, Slope and Energy features were calculated on window size = 6. Not all these statistical features significantly contribute to improving the model's performance, so we performed feature selection by fitting histogram classifier until reaching convergence. Following this, we applied permutation importance on the validation set, selecting only the features that exhibited a positive r-value. This exercise reduced the 175 statistical features to 27 features that were considered the most valuable statistical features for our analysis. As randomness is part of the algorithm, the final number of the selected statistical features might vary a little bit on run-to-run basis.
					\item 
						Next, we calculate some other clinical scoring features which has been of clinical importance and are documented in research papers such as SOFA, MEWS, qSOFA, SIRS, and ratio features such as  “HR/SBP”, “SaO2/FiO2”, and “BUN/CR” resulting in additional 7 features. 
					\item 
						Aggregate all these features to get a total of 107 features (35 original features + 35 masked features + 27 statistical features + 7 other features + 3 demographical features (age, sex, weight)) are the final candidate that goes into training along with the Labels. 
					
				\end{itemize}

	\section{Model Training}
			The dataset was split into a training set and a test set using an 80-20\% ratio. To ensure the consistency of sepsis prevalence between the two sets, special attention was paid to maintaining the same percentage of sepsis cases in both partitions.

			We employed the popular XGBoost algorithm for training the predictive model. Log loss was chosen as the loss function during the training process. Log loss is commonly used in binary classification tasks and is particularly effective when dealing with probabilistic predictions, as it penalizes both false positives and false negatives.
			
			We adopted a dynamic learning rate strategy to optimize the model during training by started with a relatively high learning rate of 0.01 and gradually decreased it as the training proceeded. This approach is known to help the model converge to an optimal solution more effectively.
			
			Training the model was a computationally intensive task, and it took approximately 4 hours on 4th gen Intel Xeon Scalable processors (codenamed Sapphire Rapids) machine. The model was trained for 3000 epochs.
			
			To address the issue of class imbalance in the dataset (which is common in medical datasets where positive cases are often rarer), we experimented with weighted log loss as an alternative loss function. However, observed that using the weighted log loss increased the false positive rate, leading to more false alarms. As false alarms could have serious implications in a medical setting, we reverted to using the regular log loss function.

					\begin{figure}
						\centering
						\includegraphics[width=\textwidth]{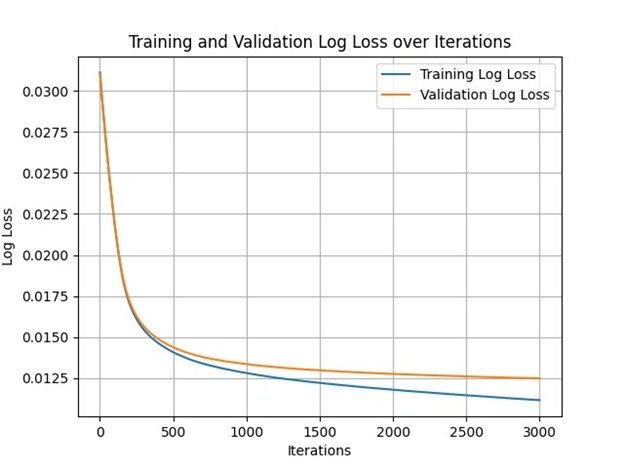}
						\caption{ Training Loss curve.}
						\label{fig:learningcurve}
					\end{figure}
			
			The training loss significantly decreased in the first 500 epochs and gradually converges around 3000 epochs (see Figure \ref{fig:learningcurve}). The validation loss is not significantly different from training loss hence we don’t see a problem of under or over fitting with validation data.
	
	\section{Results}
		\subsection{Retrospective Data Analyses}
			For evaluating the model's performance, we employed the normalized utility score, as defined by the PhysioNet Challenge. The normalized utility score is a comprehensive metric that considers various factors, including sensitivity, specificity, and prediction time. The normalized utility function provides incentives for classifiers that make timely sepsis predictions while imposing penalties for late or missed predictions and for predicting sepsis in non-sepsis cases. The scoring system is described in detail here: \cite{reyna2020early}\cite{physio-online}. 
				\begin{itemize}
					\item 
						For septic patients (at least one Labels entry of 1), classifiers predicting sepsis within the 12 hours before and 3 hours after the onset are rewarded. where the maximum reward is a parameter (1.0). Conversely, penalties are applied to classifiers failing to predict or predicting sepsis more than 12 hours before (parameter: 0.05 for very early detection, -2.0 for late detection).
					\item 
						For non-septic patients (all Labels entries of 0), we penalize the False Positives, where the maximum penalty for false alarms is a parameter (0.05; equal to the very early detection penalty). 
					\item 
						We neither reward nor penalize those that do not predict sepsis.
					
				\end{itemize}
							
			We assess the model’s performance in a clinical setting, we examined all predictions for a single encounter. For septic patients, success for the model was deemed by any score greater than the threshold within an six-hour window prior to the time of presentation. A lack of any positive predictions within this window was considered a failure. For non-septic patients, any positive predictions during the encounter were considered a failure, while a lack of any positive predictions between admission to discharge was a success. In addition to precision and recall, to determine optimal use in a clinical setting, we paid particular attention to flag rate (percentage of encounters with at least one positive flag), false positive rate (percentage of encounters with a positive prediction that did not go on to have severe sepsis), and median timeliness (In successful positive cases, median of the hours of first prediction prior to time of presentation). With respect to these metrics, a cutoff of .3 was chosen to be most useful for a live hospital setting. Other commonly used metrics such as FPR (False Positive Rate), PPV (Positive Predictive Value)  and NPV (Negative Predictive Values) are also calculated.
			
			Patient level metrics were calculated by varying the confidence threshold to see how the model performed on those confidence levels (Table \ref{tab:retrospectiveresults}). To determine the optimal performance, we tested different confidence thresholds ranging from 0.1 to 0.9. By analyzing the resulting F1 scores and normalized utility scores, we identified that a threshold of 0.3 provided the most favorable outcome. This threshold yielded a normalized utility score of 0.494.
								\begin{table}
									\centering
									\includegraphics[width=\textwidth]{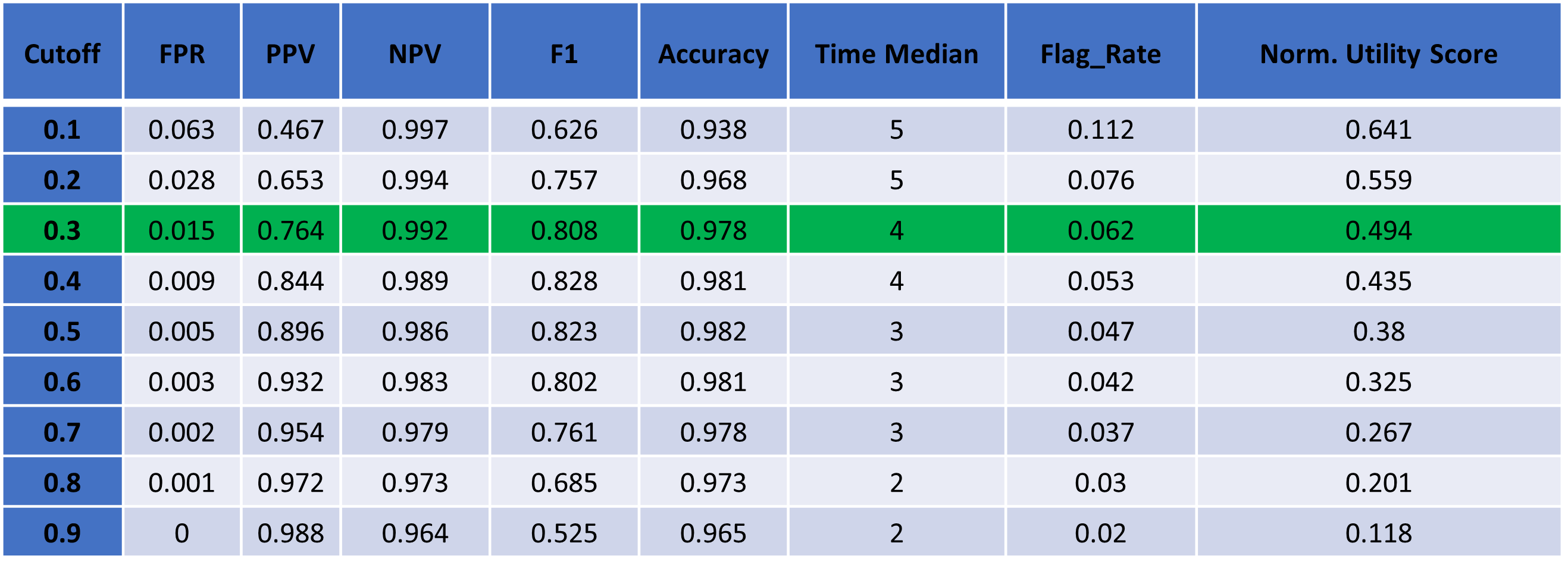}
									\caption{Different metrics on retrospective data}
									\label{tab:retrospectiveresults}
								\end{table}

		\subsection{Prospective Data Analyses}
\begin{figure}
							\centering
							\includegraphics[width=\textwidth]{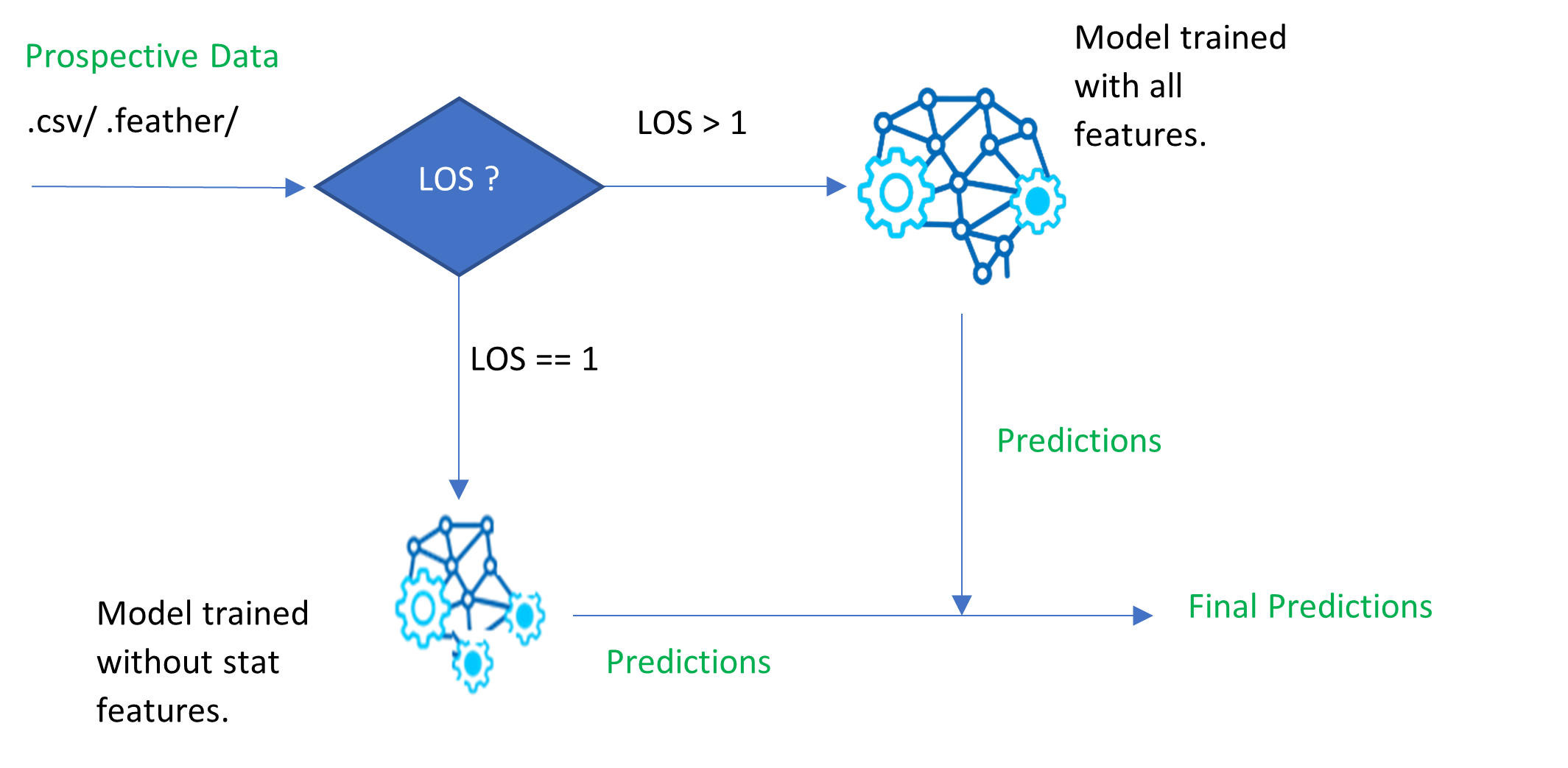}
							\caption{ Prospective Data Analyses flow}
							\label{fig:prospective_arch}
						\end{figure}

			We wanted to test model’s reliability on completely unseen data in hospital settings. We obtained first three months of hospital data from year 2023. The data consisted of total 50,014 encounters (2,030,807 hourly records) of which 6\% were septic patients. Out of this, a total of 11,479 encounters have only one hour of data. This can happen in real time scenario, hence to deal with such data we trained. As statistical features based on windowing cannot be obtained for encounters with just one hour of data, we trained an additional XGBoost model only on non-statical features. The remaining data points are sent through the model trained with all the features. In the end we combine all the results together and run our row-level and patient level metrics. See the architecture in \ref{fig:prospective_arch}.
		
			On prospective data, the normalized utility score at 0.378, lower than that of the test data. Similarly the F1 was 67.1\%.  Despite using this ensembling technique we could see the model performance dropped by 10\%, the reason being nearly 40\% of the prospective data has only patient record values with less than 5 hours, hence the stats features  for these patients couldn’t be calculated, as our early prediction window is 6 hours. Since stats features turn out to be important features for sepsis prediction, the model performs poorly in these cases.  \ref{tab:prospectiveresults}.  
						\begin{table}
							\centering
							\includegraphics[width=\textwidth]{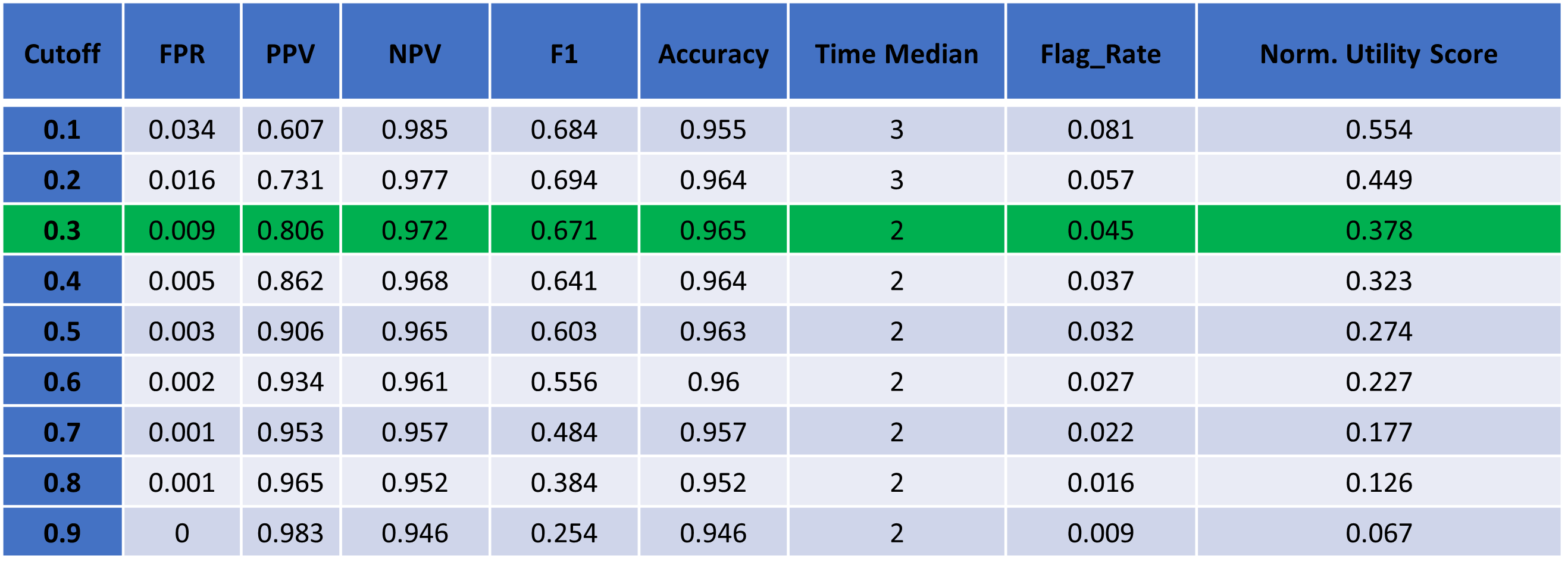}
							\caption{Different metrics on prospective data}
							\label{tab:prospectiveresults}
						\end{table}

	\section{Explainable AI using SHAP}
		\label{sec:explainableai}
\begin{figure}
						\centering
						\includegraphics[width=\textwidth]{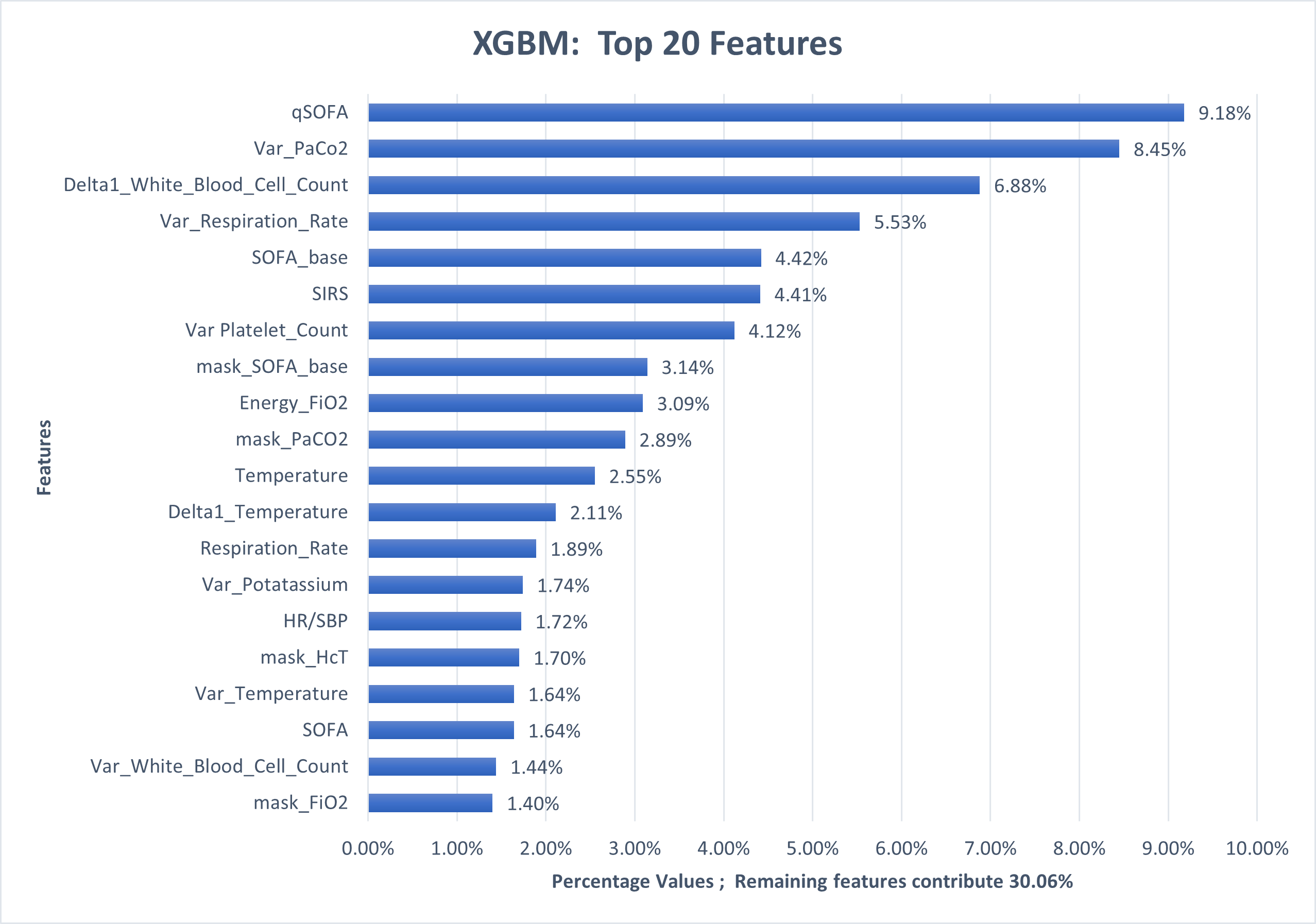}
						\caption{Efficient AI}
						\label{fig:featureimp}
					\end{figure}

		SHAP (Shapley Additive Explanations) analysis is a powerful technique used in machine learning for interpreting and understanding the predictions made by complex models. Shapley values are calculated relative to a baseline value. This baseline typically represents the expected or average prediction of the model. It provides insights into the contribution of each feature in the prediction of a specific outcome. We used SHAP for better interpretation of the model and plotted the feature importance plot (as the percentage of the contribution of individual feature) of the top 20 features that contribute to a model’s prediction (See Figure \ref{fig:featureimp})). We observe that the contribution of top 20 features in the prediction of sepsis is almost 69.9\%. The remaining 87 features contribute to 30.06\%. In our analyses, qSOFA, variance of the PaCO2, SOFA, Delta of WBC, Variation in Respiration, SOFA, SIRS and others to be the most important features. 
	
		We looked at some of the scholarly articles to see how the top features are associated with the sepsis. 
		\begin{itemize}
			\item 
			qSOFA (quick Sequential Organ Failure Assessment): The qSOFA is used for the rapid assessment of patients suspected of having sepsis  \cite{singer2016third}. Freund el. al. \cite{Freund2016} concluded that the use of qSOFA resulted in greater prognostic accuracy for in-hospital mortality than either SIRS or SOFA, reflected in this plot as well. A similar related study by Fernando et al. \cite{fernando2018} shown qSOFA to have poor sensitivity but higher specificity compared to SIRS. 
			
			\item
			Var\_PaCO2:  Monitoring PaCO2 and repsiratory rate is crucial in the care of septic patients to assess respiratory function and guide therapeutic interventions \cite{nasa2012severe}. Abnormal PaCO2 levels can provide important information about the severity of respiratory compromise and the need for respiratory support. 
			
			\item
			Delta1\_White\_Blood\_Cell\_count: Delta1\_White\_Blood\_Cell\_count is the difference in the WBC count of current hour and previous hour. The WBC count is an important component to calculate the SIRS score. The changes in WBC count, body temperature, and heart rate reflect inflammation and the host response to infection \cite{singer2016third}. 

			\item 
			Var\_Respiration\_Rate: Respiration rate is an important constituent for calculation of SOFA, qSOFA, SIRS and MEWS. It has been included in the assessment of sepsis according to the Sepsis-3 criteria and other relevant studies \cite{seymour2016assessment} \cite{Freund2016} \cite{churpek2017quick}. Monitoring respiration rate is crucial in identifying potential respiratory dysfunction associated with sepsis and contributes to the early prediction and management of septic patients. Our model picked up the variance in Respiration rate as the 4th most important feature for sepsis prediction.   
			
			\item 
			SOFA\_base (Sequential Organ Failure Assessment) score: The SOFA score serves as a valuable tool for describing and predicting organ dysfunction in sepsis patients, as validated by studies such as \cite{vincent1996sofa} \cite{seymour2016assessment} and \cite{ferreira2001serial}, supporting its widespread adoption in sepsis detection and prognosis. Ferreira et al. \cite{ferreira2001serial} concluded that SOFA scores are reliable in evaluating organ dysfunction in the initial days of ICU admission serves as a valuable prognostic indicator. Regardless of the initial score, a rise in SOFA score within the first 48 hours in the ICU is indicative of a mortality rate of at least 50\% \cite{ferreira2001serial}. 

			\item 
			SIRS (Systemic Inflammatory Response Syndrome): SIRS is a generalized inflammatory response, and its progression to sepsis involves evidence of infection, with early recognition crucial for timely intervention \cite{bone1992definitions}. SIRS is a part of historical context and is now replaced by SOFA. Zhang et al. shown that the SIRS criteria are weaker than the SOFA criteria with respect to their predictive efficacy in hospital mortality \cite{zhang2019systemic}. 
			
			\item 
			Platelet count: Platelet count is an important hematological parameter in the context of sepsis, as it can reflect the body's response to infection and inflammation.  It has been included in the assessment of sepsis to calculate SOFA score, according to the Sepsis-3 criteria \cite{seymour2016assessment} \cite{Freund2016}. Schupp et al. provided evidence that the platelet count represents a reliable tool for the diagnosis of sepsis and can be effectively used for prognosis of septic shock during the first 10 days of ICU hospitalization \cite{schupp2023diagnostic}.
			
			\item 
			mask\_SOFA\_base: masked feature corresponding to the SOFA.
			
			\item 
			Temperature and Delta1\_Temperature: Monitoring body temperature is essential for recognizing signs of infection and systemic inflammatory response, which are crucial components of sepsis detection. The body temperature is also used to calculate SIRS and MEWS scores. In a study conducted by Khodorkovsky et al. \cite{KHODORKOVSKY2018372}, in emergency department patients, hypothermia was correlated with increased time to initial antibiotics, length of stay, rate of ICU admission, and mortality. Another study by Doman et al. \cite{Doman2023}, shown in a controlled study that the use of temperature does not improve the sepsis prognosis. In our experiment, the predictive contribution of Temperature and Delta1\_Temperature, collectively, is 4.66\%.
		
			\item 
			Respiration\_Rate: See Var\_Respiration\_Rate above. 
			
			\item 
			Var\_Potassium: Potassium imbalance may impact organ functionality, therefore managing Potassium level is crucial in sepsis management. Association between Potassium and prognosis of sepsis has been inadequately investigated. Some machine learning studies found these two to be correlated. Wang et al. \cite{wang2021model} found potassium to be the 9th most important feature among a total 55 features in their data and it is closely related to the occurrence of sepsis. Increasing evidence suggests potassium channels play a role in sepsis-related cardiovascular dysfunction, systemic inflammation, and organ damage, potentially influencing sepsis development post-infection \cite{sordi2011early}. Other studies based on ML algorithms also show the importance of Potassium in sepsis prediction \cite{yue2022machine} \cite{zhao2021early}.
			\item 
			HR/SBP (Shock Index): The sepsis shock index is a hemodynamic parameter calculated as the ratio of  heart rate (HR) and systolic blood pressure (SBP). A higher sepsis shock index may indicate more severe cardiovascular dysfunction and can be indicative of a worse prognosis in sepsis patients. The HR and/or SBP are used to calculate the early sepsis warning indicators such as MEWS, SIRS and qSOFA. A number of studies found shock index among the most important features \cite{morrill2019signature} \cite{li2019tasp} \cite{xie2023predicting}.
			
			\item 
			mask\_HcT: As mentioned above, masking is an important feature as they lead to creating additional features that may help retain vauable information about the frequency of data collection. The masked feature of HcT is one of the important features in sepsis prediction. According to Luo et al. \cite{luo2022association}, The low HcT level is an independent risk factor for the increase of the 30-day mortality in patients with sepsis and can be used as a significant predictor of the clinical outcome of sepsis.   
			Monitoring hematocrit levels is crucial in sepsis prediction as it provides insights into oxygen transport, guiding interventions to optimize tissue perfusion and oxygenation. 			
			
			\item 
			Var\_Temperature: see Temperature
			\item
			SOFA: see SOFA\_base
			\item 
			Var\_White\_Blood\_Cell\_Count: see Delta1\_White\_Blood\_Cell\_count.			
			\item 
			mask\_FiO2: FiO2, or Fraction of Inspired Oxygen, is the percentage of oxygen in the air a person breathes. In sepsis detection, monitoring and adjusting FiO2 levels are crucial to ensure optimal oxygenation, addressing potential tissue hypoxia in septic patients. Chicco et al., when analyzing septic shock data, found the FiO2 to be the 5th most important feature when created a model using RandomForest \cite{chicco2021data}. 
			
		\end{itemize}

\section{Conclusion and Future Work}
	In conclusion, this study presents a significant advancement in the early detection of sepsis using Machine Learning models. Sepsis is a life-threatening condition that requires prompt timely intervention for improved patient outcomes, and the absence of a single specific test for sepsis detection makes early recognition challenging. This research demonstrates the potential of leveraging clinical data from Montefiore Medical Center to predict the onset of sepsis up to 6 hours in advance. Early detection of sepsis is paramount in mitigating mortality rates, and our approach offers a valuable tool for timely intervention. By leveraging machine learning techniques, we have shown that it is possible to harness the wealth of clinical data available to enhance sepsis prediction, ultimately leading to improved patient outcomes.

	The XGBoost model, trained on a rich dataset obtained at Montefiore Medical Center in Bronx, NY, USA,  showcases commendable predictive capabilities. The model achieved a normalized utility score of 0.494 on test data and 0.378 on prospective data with threshold of 0.3, demonstrating its ability to identify sepsis at an early stage accurately. Moreover, the model attained an impressive F1 score of 80.\% for test data and 67.1\ for the same threshold, highlighting its potential to be integrated into clinical decision-making processes effectively. Overall, this ML/DL technique enable us to use a collection of different scoring techniques, lab values, demographic information and derived features to help prognosis of sepsis. 
	
	Moving forward, our work paves the way for further refinement and integration of machine learning models in clinical settings. We are also exploring to extend this solution to multimodal approach by ensembling an XGBoost model and NLP model trained with clinical notes. Continued research and collaboration in this field hold the promise of revolutionizing early sepsis detection and significantly advancing patient care in critical healthcare scenarios.

\section{Disclaimer}
The information presented in this research paper is intended for educational and informational purposes only and should not be regarded as a substitute for professional medical advice, diagnosis, or treatment. It relies on historical data for training and may not accurately reflect individual cases in real-world scenarios. It is crucial to consult a qualified healthcare professional for precise diagnosis and treatment decisions. The model's predictions serve as supplementary information and should not be considered a definitive diagnosis. Users should be aware of the potential risks associated with misinterpretation or misuse of the model's output. The developers and creators disclaim any liability for damages arising from its use, and users assume full responsibility for the interpretation and application of the model's predictions. Ethical considerations, adherence to patient privacy, and compliance with legal requirements are essential when utilizing this sepsis detection method and model. By using the method and model, users acknowledge and agree to the terms and conditions outlined in this disclaimer.



\end{document}